# An Improved Approach for Cardiac MRI Segmentation based on 3D UNet Combined with Papillary Muscle Exclusion

Narjes Benameur[1], Ramzi Mahmoudi[2], Mohamed Deriche[3], Amira fayouka[2], Imene Masmoudi[4], Nessrine Zoghlami[5]

1 Higher Institute of Medical Technologies of Tunis, Laboratory of Biophysics and Medical Technologies, University of Tunis El Manar, Tunisia narjes.benameur@istmt.utm.tn
(Corresponding author)

2 Faculty of Medicine of Monastir, Medical Imaging Technology Lab - LTIM-LR12ES06, University of Monastir, Tunisia.
ramzi.mahmoudi@esiee.fr

3 Artificial Intelligence Research Centre, Ajman University, United Arab Emirates.
m.deriche@ajman.ac.ae

4 University of Tunis El Manar, National Engineering School of Tunis, Tunisia
imen.masmoudi@etudiant-enit.utm.tn

5 University of Tunis El Manar, IPEIEM, 10587 Tunis, Tunisia
nesrine.zoghlami@ipeiem.utm.tn

**Abstract:**
***Background***: Left ventricular ejection fraction (LVEF) is the most important clinical parameter of cardiovascular function. The accuracy in estimating this parameter is highly dependent upon the precise segmentation of the left ventricle (LV) structure at the end diastole and systole phases. Therefore, it is crucial to develop robust algorithms for the precise segmentation of the heart structure during different phases.
***Methodology***: In this work, an improved 3D UNet model is introduced to segment the myocardium and LV, while excluding papillary muscles, as per the recommendation of the Society for Cardiovascular Magnetic Resonance. For the practical testing of the proposed framework, a total of 8,400 cardiac MRI images were collected and analysed from the military hospital in Tunis (HMPIT), as well as the popular ACDC public dataset. As performance metrics, we used the Dice coefficient and the F1 score for validation/testing of the LV and the myocardium segmentation.
***Results***: The data was split into 70%, 10%, and 20% for training, validation, and testing, respectively. It is worth noting that the proposed segmentation model was tested across three axis views: basal, medio basal and apical at two different cardiac phases: end diastole and end systole instances. The experimental results showed a Dice index of 0.965 and 0.945, and an F1 score of 0.801 and 0.799, at the end diastolic and systolic phases, respectively. Additionally, clinical evaluation outcomes revealed a significant difference in the *LVEF and other* clinical parameters when the papillary muscles were included or excluded.
***Conclusion***: The proposed framework is shown to outperform state-of-the-art methods by around 0.1 in terms of dice index, demonstrating its accuracy in the precise assessment of the left ventricular function.

## 1. INTRODUCTION

Ventricular function evaluation in routine clinical practice has seen significant development in recent years, thanks to the technological revolution we are experiencing across various fields, including medical imaging. In addition to the development of a helium-free MRI machine and the latest trends of using virtual sequences derived from artificial intelligence [1,2], radiologists have also benefited from the recently development to streamline protocols for interpreting MRI images [3].

The protocol of LV function often begins with a single shot fast spin echo, gradient echo or bright blood balanced steady state free precession (bSSFP) images acquired in 3 planes of views: 2-chamber, 4-chamber, and short axis. Then, endocardial and epicardial contours were delineated by radiologist using a series of images that represents the entire cardiac cycle. The Left ventricular ejection fraction (LVEF) and other clinical parameters such as end diastolic volume, end systolic volume and myocardial mass are computed using the software available on the console acquisition [4,5]. Therefore, the specificity of these clinical parameters is highly influenced by the accuracy of LV segmentation [6]. The analysis of these images requires more than 10 min per patient for an experienced radiologist. Moreover, In the absence of automated segmentation technique, the delineation of epicardial boundary can be difficult to detect in the presence of papillary muscles [7].

The segmentation of the endocardium and epicardium of the left ventricle (LV) as well as the right ventricle (RV) from magnetic resonance imaging datasets has received significant attention from researchers in recent years. Several studies have been developed to create state-of-the-art methods in semi-automatic cardiac segmentation [8-10]. These works have provided ground truth contours and a set of evaluation metrics to compare different approaches. Moreover, more efforts have been made to automate the segmentation of various anatomical structures and lesion detection. Non-learning-based algorithms such as statistical shape modelling, level sets, active contours, multiple atlases, and graphical models have shown promising results on limited datasets [11-13]. Some of these techniques heavily rely on handcrafted features and thus require domain knowledge and expert intervention. To overcome these limitations, methods based on CNNs have been adopted for a variety of computer vision and pattern recognition tasks. The most common ones are image classification [14-16] and semantic segmentation using Fully Convolutional Networks (FCNs) [17]. Among all these applications, deep Convolutional Neural Networks (CNNs) have demonstrated greater representation capability and hierarchical learning. Recently, in MRI, FCNs and their popular extensions like U-NET (convolutional neural networks for biomedical image segmentation) [18,19] have achieved remarkable success in segmenting various heart structures for automating cardiac diagnosis from MRI images [20]. Moreover, with the availability of huge amounts of labeled data and increased computing power, information technology is being increasingly utilized [21]. With more powerful and general-purpose Graphics Processing Units (GPUs), CNN-based methods have the potential to be applied in daily clinical practice.

In recent works, architectures based on deep learning have shown greater modelling capacity and the ability to learn highly complex features. However, most of the proposed models detect the endocardial contour along with the papillary muscles. These structures have a very similar signal intensity level to the myocardium in all MRI sequences and during systole, they contract before the left ventricle [22]. The latest recommendation, from the society for cardiovascular Magnetic resonance (SCMRI), suggests excluding papillary muscles from the LV volume [23]. In line with this context, some researches proved that including the papillary muscles and trabeculae to left ventricle significantly affects Left ventricle end diastolic volume (LVEDV) and left ventricular mass [24,25]. More recently, Kim el al. [26] reported that the inclusion of papillary muscles during the segmentation process increased the total of LVEDV. In their study they found that the contribution of papillary muscle to the total left ventricle volume was 11.9 % while it accounted for 20.2 % of the left ventricular mass. Similarly, another study showed a notable difference in LV measurements in patients with hypertrophic cardiomyopathy depending on whether papillary muscles were included or excluded from the blood cavity [27].

In this paper, we propose a new approach of myocardial and LV segmentation without papillary muscles based on an improved 3D UNet that reaches competitive results compared to other models. The main contributions of this study are:

1. We develop a deep learning framework for the automated segmentation of myocardium and left ventricle, with excluding papillary muscles, in accordance with the last recommendation of Society for Cardiovascular Magnetic Resonance.
2. Segmentation results were tested on two medical images datasets (public and internal datasets), which prove the accuracy of our model.
3. A large dataset was collected for this study that include 7,800 MRI images related to 156 subjects. The proposed dataset will serve as a valuable source of cardiac data.
4. We study the difference between LVEF measurements derived from LV segmentation with and without papillary muscles.

The remainder of the paper is structured as follows: Section 2 provides an overview of the related LV segmentation methods. Section 3 presents a brief description of the used datasets. Section 4 gives a detailed description of the methodology of our deep learning model. Section 5 showed the evaluation measures used in this study and details the experimental results and analysis. The discussion of the main results is presented in section 6. Finally, we conclude the paper in Section 7.

## 2. Related works

In the last two decades, a wide range of approaches have been proposed to segment the left ventricle using MRI. The initial works began with the use of classical techniques such as the thresholding and the clustering approaches [28]. Furthermore, many researches have utilized cardiac atlas and deformable models for heart segmentation [29]. However, most of these techniques require user interaction and a tedious learning phase, and their segmentation results are influenced by the gray level variations around the myocardium. As a result, their outcomes are prone to numerous errors. With the rise of deep learning, advanced techniques have shown promising results in cardiac segmentation, However, new challenges have emerged in achieving satisfactory accuracy.

In this regard, numerous studies have been conducted, presenting powerful algorithms that have yielded good results. For instance, the study conducted by Brahim et al. [30] proposed the SegU-Net model, which combines the U-Net and Seg-Net architectures. In another study Li et al. [31] used a U-Net encoder combined with a ConvLSTM decoder for segmentation. Similarly, Hasan et al. [32] presented the CondenseUNetis model, integrating the CondenseNet12 and U-Net architecture. These works have demonstrated that the use of the U-Net architecture can significantly enhance the accuracy of myocardial segmentation.

Many studies have also adopted the U-Net architecture for myocardial segmentation. The researchers mentioned in references [33-35] exploited the 3D version of U-Net, while others [36, 37] preferred using the standard U-Net. These studies have clearly demonstrated the advantages of the U-Net architecture in terms of precise myocardial segmentation, thereby contributing to improving the process of diagnosing cardiac diseases by accurately detecting myocardial contours and classifying diseases more precisely and efficiently.

Furthermore, there have been several other investigations conducted on the segmentation of the left ventricle. For example, Yan et al. [38] employed an improved version of the SegNet model and achieved a Dice similarity score (DSC) of 0.878. Zhuang et al. [39] utilized the YOLOv3 model, yielding a DSC of 0.93 Additionally, a separate study [40] proposed the Light U-Net model, which demonstrated a remarkable DSC of 0.971. These studies highlight the efficiency of different models in segmenting the left ventricle, thus contributing to the progress of segmentation techniques in this particular domain.

In the last years, with the publication of the latest recommendation by the American College of Cardiology, some authors have emphasized the importance of excluding papillary muscles in the left ventricle segmentation process. The study conducted by Awasthi et al. [41] specifically addressed this issue and demonstrated the significance of their exclusion to achieve precise and high-performing results. The outcomes of this study revealed a significant improvement in the performance of the LVNet model compared to UNet, particularly when the papillary muscles were excluded. Another recent study [42] also showed that the elimination of pillars facilitated the interpretation of the middle and apical sections and contributed to the localisation of pathologies and the estimation of clinical parameters.

These findings support the notion that the exclusion of papillary muscles is a crucial element in obtaining more precise and high-performing results in left ventricle segmentation. By eliminating these structures in the segmentation process, the LVNet and the UNet model can focus more on accurately segmenting the left ventricle itself, thereby enhancing the overall performance of the model.

### 3. MRI Dataset description

Two datasets were used in this study: the first is a local dataset collected from the Military Hospital of of Tunis Tunisia (HMPIT) which we called "MHT-MRI". It comprises DICOM cine MRI images representing image slices related to 156 subjects, including 7,800 MRI images (50 images per subjects * 156 subjects) of patients with cardiac diseases and healthy controls. The acquired data were fully anonymized and processed in compliance with regulations established by the local ethics committee of the HMPIT hospital of Tunisia. The data is available from the corresponding author upon request. All clinical characteristics of the studied population are summarized in table 2.

**Table 2**. Clinical characteristics of the studied population

| Characteristics | Subjects with cardiac complications (n=90) | Healthy controls (n=66) |
| --- | --- | --- |
| *Age (years)* | 47 ± 12 | 49 ± 13 |
| *Sex (male /female)* | 49/41 | 31/35 |
| *Temperature (°c)* | 36.8 [36.4-39,1] | 37,1 [36.9-40.2] |
| *Pulse (/min)* | 78 [75-89] | 76 [74-90] |
| *History of CAD* | n= 20 (22.22%) | n= 17 (25.75%) |
| *Hypertension* | n= 22 (24.44%) | n= 12 (18.18 %) |
| *Type 1 - Diabetes* | n= 8 (8.88%) | n= 13 (19.69%) |
| *Type 2 - Diabetes* | n=40 (44,44%) | n= 24 (36,36%) |
| *Type of cardiac complications:* | | Myocarditis |
| | | Myocardial Infarction |

The public ACDC dataset were also used in this study [43]. It comprises clinical examinations conducted at the University Hospital of Dijon. This database consists of a training set of 100 cases and a test set of 50 cases. Cine MRI images were acquired using a standard SSFP sequence. Most cases contain approximately 10 series of short-axis MRI images. The database comprises 4 classes of pathologies: myocardial infarction (MINF), dilated cardiomyopathy (DCM), hypertrophic cardiomyopathy (HCM), abnormal right ventricle (RVA), and a group of normal subjects. Within the same database, manual segmentation of the left ventricular cavity and right ventricular cavity is provided as ground truth for all images at two specific time points: end-diastole (ED) and end-systole (ES).

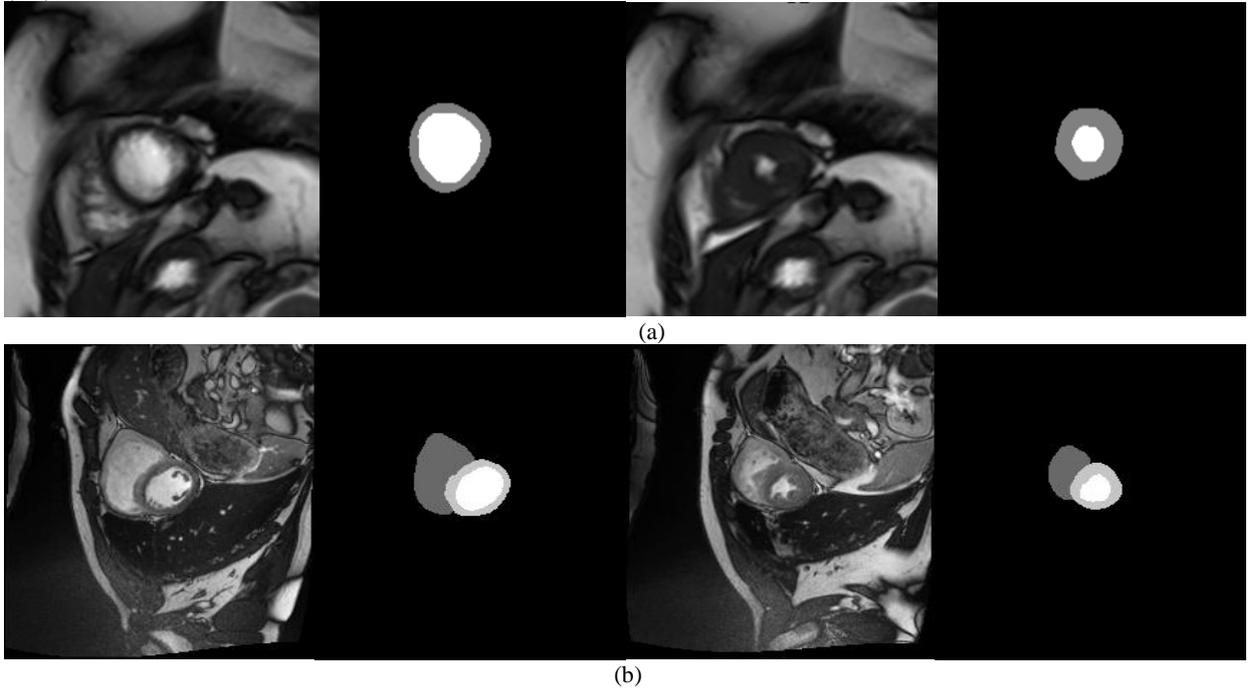

*Figure 1: Illustration of a typical cine MRI images;) MRI images in short axis view from the local dataset (a) and from the ACDC dataset (b) at end-diastolic (ED) and at end systolic (ES) frames with masks.*

**Table 3.** Database repartition.

|  | *Training Set (70%)* | *Test Set (20%)* | *Validation Set (10%)* | *Total* |
|---|---|---|---|---|
| *Number of images* | 5,880 | 1,680 | 840 | 8,400 |

## 4. Proposed Framework
### 4.1. Preprocessing

- *Resizing MRI images and their masks*

To input data into our model, we standardized the shape of all images. A bounding box was utilized to localize the heart and determine its center's coordinates in space. Subsequently, based on the center's position relative to the desired shape (156, 156, 6), we extracted the portion of the image containing the heart. If the image's dimensions were less than 156 for either width and/or height, a new image was created from the original one using a matrix with the desired shape, prefilled with zeros. Figure 2 illustrates the resulting image after resizing.

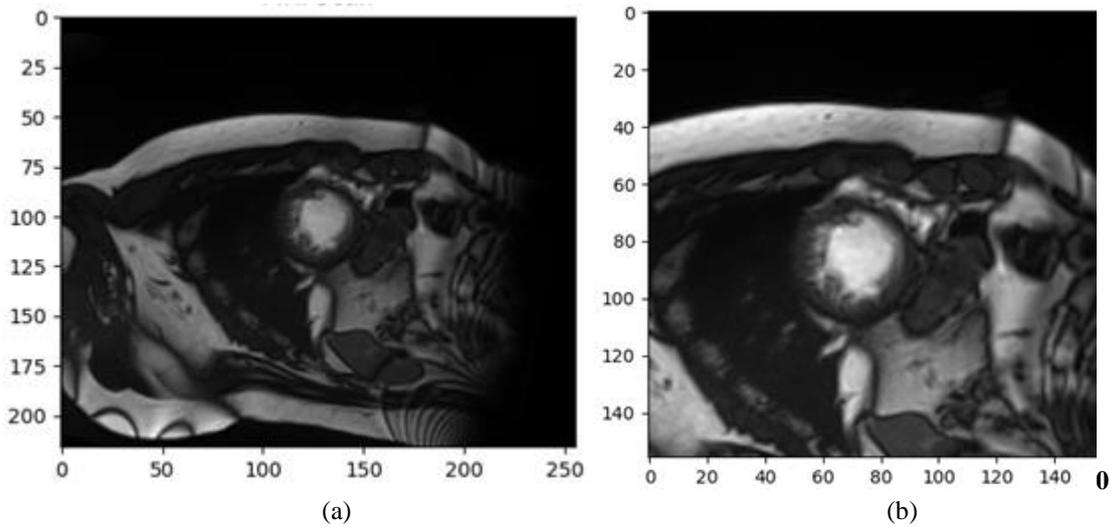

*Figure 2. Resizing step (a) MRI image before (a) and after (b) resizing process.*

- *Cleaning masks*

The masks generated from segmentation results available in public datasets play an important role in evaluating the performance of deep learning models. Since the aim of this study is to segment left ventricle and myocardial while excluding the papillary muscles, a mask-cleaning step is required to remove unwanted myocardial structures from different images. Accordingly, two radiologists (7 years, and 12 years of CMRI experience, respectively) participated in identifying the left ventricle without papillary muscles using the CVI42 (Circle Cardiovascular Imaging) software available in the acquisition console. Based on these results, new masks were generated for both datasets to train the proposed 3D UNet.

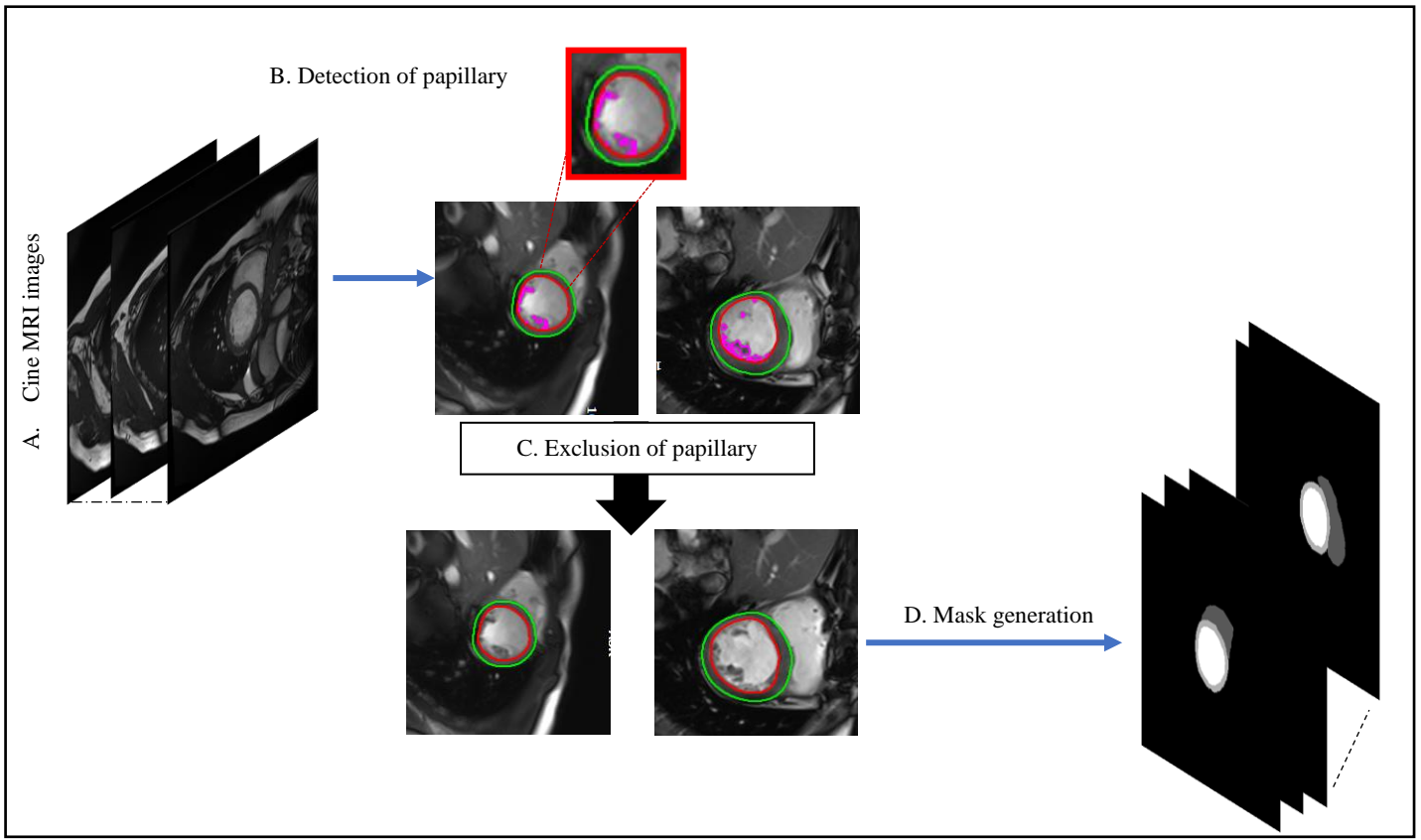

*Figure 3. Cleaning masks process.*

- *Patching the images into smaller cubes*

Instead of using the entire image, a patch-based technique was used. This includes randomly cropping small patches from the image and feeding them into the fitting process. Our model's architecture requires that the input shape (x, y, z) follow these constraints:
- x, y, and z are a multiple of $2^2$
- x, y, and z are greater or equal to 4

in our study, the patching technique was used to extract 64 x 64 x 4 volumetric medical images.

### 4.2. *Architecture of the modified model 3D-Unet*

The proposed model architecture represents a 3D U-Net designed for medical image segmentation tasks. Here are some key features of our approach:

- Encoder-Decoder Structure: Similar to the original U-Net, the model consists of contracting (encoder) and expanding (decoder) paths, creating a U-shaped architecture.
- Contextual Feature Extraction: The encoder utilizes a series of 3D convolutional layers, each followed by Batch Normalization and ReLU activation, to progressively extract and downsample features, capturing contextual information at different scales. Max Pooling layers further contribute to downsampling and reducing spatial dimensions.
- Precise Localization: The decoder path employs 3D transposed convolutions for upsampling the feature maps, effectively increasing their resolution. Additional 3D convolutions are used to refine the feature maps for accurate localization of structures within the image.
- Skip Connections: The model incorporates skip connections that bridge corresponding levels of the encoder and decoder paths. These connections facilitate the transfer of fine-grained details from the encoder to the decoder, leading to more precise segmentation results.
- Output Layer: The final layer of the model is a 3D convolutional layer with a number of filters equal to the number of target classes. This layer produces the segmentation map, which has the same spatial dimensions as the input image.

The model is trained on 3D patches of size 64 x 64 x 4 voxels extracted from volumetric MRI images. This information is explicitly shown in the input layer of the provided architecture diagram. The training process involves feeding the model the 3D patches and optimizing its parameters to minimize a chosen loss function: Dice loss. The architecture's strengths lie in its ability to capture multi-scale contextual information, effectively localize structures, and produce accurate segmentation maps, making it a valuable tool in medical image analysis. Figure 4 illustrates the architecture of the proposed model for the segmentation of myocardium and left ventricle:

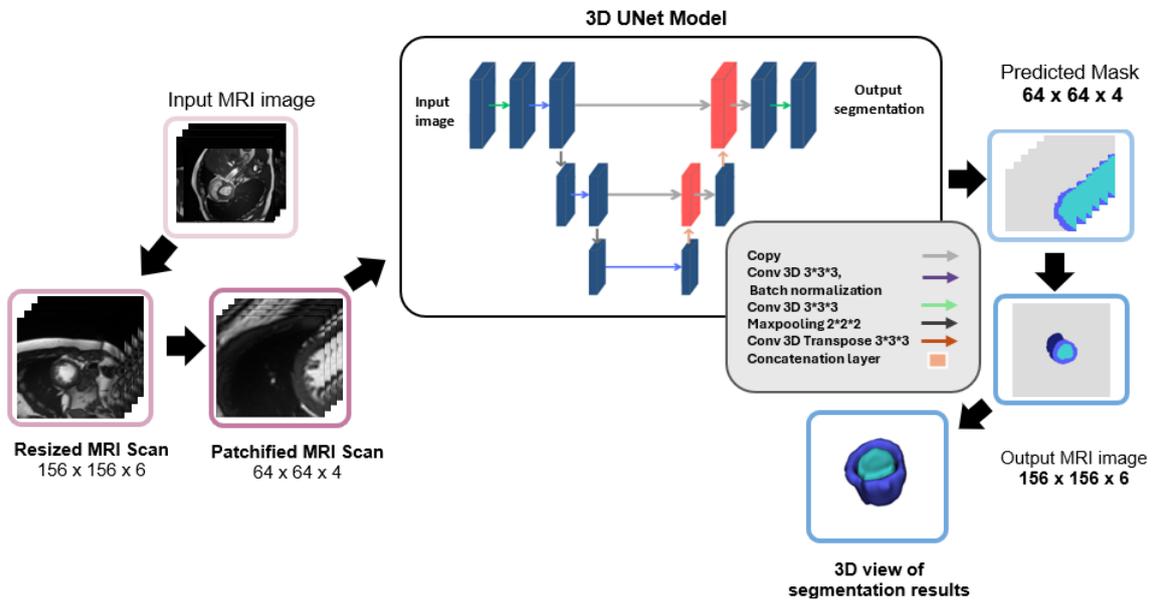

**Figure 4**: *The proposed myocardial segmentation architecture.*

### 4.3. *Hardware Setup for Deep Learning Network Training and Trials*

The proposed architecture was trained for 100 epochs in a Google Colab Pro Environment having 61 System RAM using a batch size of 32 and the ADAM optimizer with a learning rate that evolved from 0.005 for the first 40 epochs, to 0.001 between the 41st and 60th epochs and gradually decreased to reach a value of 0.0004457 using a callback. The model was implemented in Python 3.10.12 using Keras 3.3.0 deep learning library, with TensorFlow 2.13.0 as the backend and Nvidia-SMI L4 Ti having 22.5 GB graphics memory. All hyperparameters used in this study are indicated in Table 4:

**Table 4.** Hyper-parameters and their values.

| Hyper-parameters | Values |
|---|---|
| Activation functions | Relu, sigmoid |
| Learning Rate | [0.005, 0.001 .. 0.0004457] |
| Batch Size | 32 |
| Epochs | 100 |
| Optimizer | ADAM |
| Padding | same |

*4.4. 3D Reconstruction of segmentation results*

3D LV and myocardium geometries were generated from our segmentation output using CATIA V5 (Dassault Systems) software. The algorithm takes the output results as the input. During the 3D reconstruction, the results of LV contours segmentation with 3D UNet at end diastolic and end systolic phases in short axis view were used to reconstruct the myocardium and the LV surface. A workflow summarizing the different processing steps is shown in figure 5:

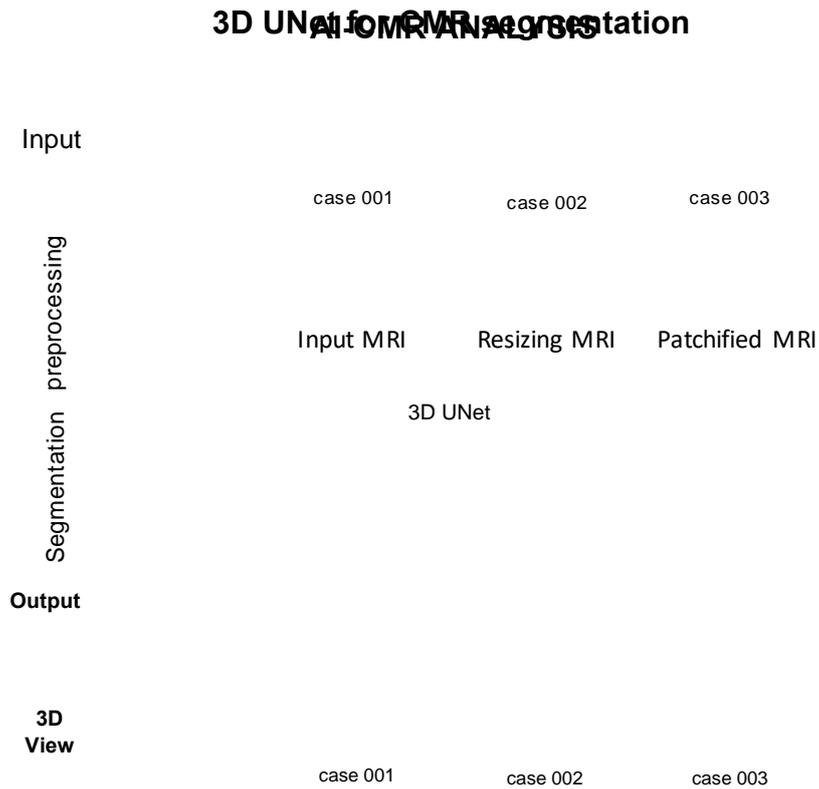

*Figure 5. A description of the overall steps of our segmentation network.*

*4.5. Validation and Statistical Analysis*

Dice coefficient, Dice loss, F1 score and Intersection over Union (IoU) were used to evaluate segmentation accuracy and to assess precision between the obtained results from DL approach and the ground truth segmentation performed by radiologists. These metrics were reported for both LV and myocardium geometries.

To examine the impact of including or excluding papillary muscles on clinical parameter measurements, Bland-Altman plots were generated for the following parameters: EDV, ESV, myocardial mass and LVEF. Additionally, the segmentation performance obtained from our proposed approach was compared with results from other deep learning approaches that utilized the same public dataset for training. The results are expressed as mean ± standard deviation (SD). All statistical analysis was performed using IBM-SPSS Statistics (Windows, version 27.0.1).

## 5. Results

### 5.1. Evaluation Metrics

In our study, we used the Dice Similarity Coefficient (DSC) as a metric to assess the overall performance of our MRI images segmentation approach. The DSC is widely recognized and used in the discipline to quantify the similarity among units of contours or masks. The computation of the DSC is primarily based on the following equation:

$$\text{DSC} = 2 \frac{|P \cap G|}{|P| + |G|} \qquad (1)$$

where P is the reference contour or masks, and G is the contour or masks expected with the aid of using our approach. The numerator is the dimensions of the intersection among the 2 units, at the same time as the denominator is the sum of the sizes of the 2 units.

A DSC near 1 shows a correct segmentation with an excessive correspondence among the expected contours and the reference contours. On the other hand, a DSC near zero shows an erroneous segmentation or a negative correspondence among the contours, indicating much less correct results.

The F1 score is a metric utilized in our paper to assess the overall performance of our segmentation method. It combines each precision and recall offering a standard degree of the overall performance of the model. The calculation of the F1-score is presented as:

$$\text{F1 score} = 2 * (\text{precision} * \text{recall}) / (\text{precision} + \text{recall}) \qquad (2)$$

in which precision represents the capacity of our technique to successfully discover superb pixels, and recall represents the capacity of our technique to locate all superb pixels. An excessive F1 score suggests a correct segmentation with a good aggregate of precision and recall while a low F1 score indicates erroneous segmentation or a poor balance of precision and recall.

we also assess the degree of similarity between the actual segmentation and the anticipated segmentation using other statistical measures, such as the Jaccard Coefficient, also referred to as Intersection-Over-Union (IoU) as shown in Eq. 4:

$$\text{IoU} (\%) = TP/(TP + FP + FN) \times 100 \qquad (3)$$

where FP and FN are the false-positive and false-negative values, respectively, and TP and TN are the true-positive and true-negative numbers.

### 5.2. Segmentation results

Figure 6 shows the performance of the proposed model in terms of dice measure, loss, IoU and F1 score. The green and red show the training and validating values of LV segmentation results from the 3D UNet model. As shown in the figure. At first, the variation was relatively significant but starting from epoch 40, it tends to be smaller, demonstrating an improvement in the validation stability of our model.

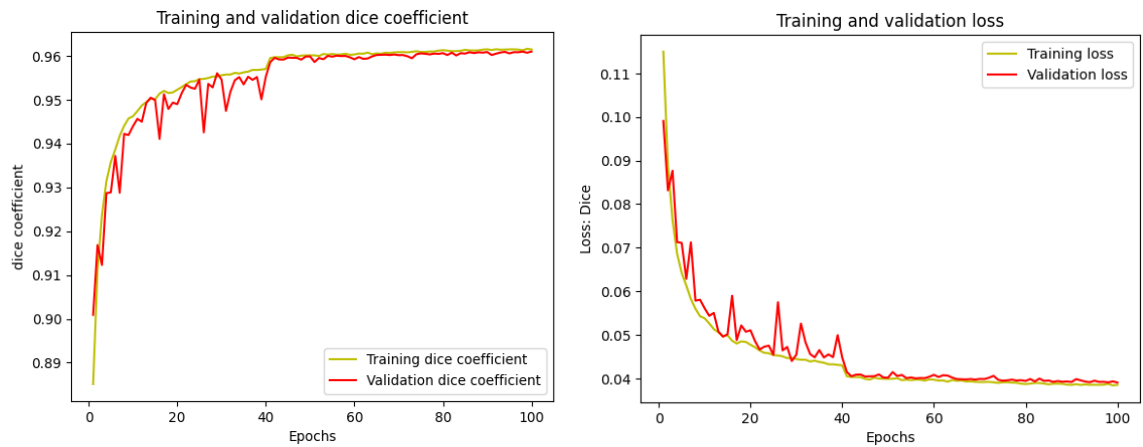

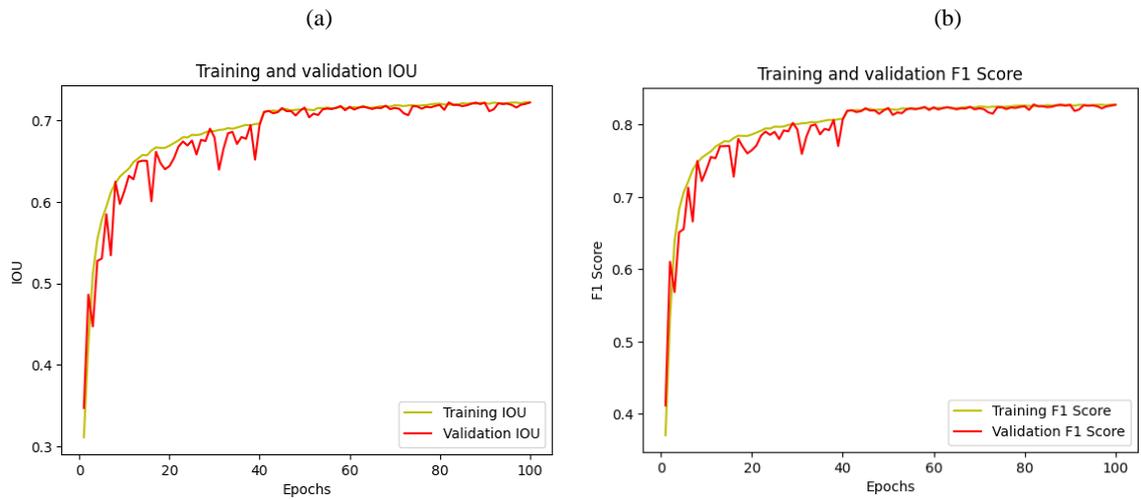

res: (a) urves

derived from our improved 3D UNet can be quantitatively assessed in figure 7. As shown in figure, the geometry of the LV and myocardium without papillary muscles can easily observed from the output images. Furthermore, The LV and myocardium structures are very similar to those obtained from the ground truth. The LV is clearly identified without papillary muscles which highlight the accuracy of the proposed approach. We can also observe that the proposed 3D UNet is able to capture the LV and myocardium structures from different slices: basal, medio basal and apical slices, demonstrating the robustness of the model against low tissue contrast.

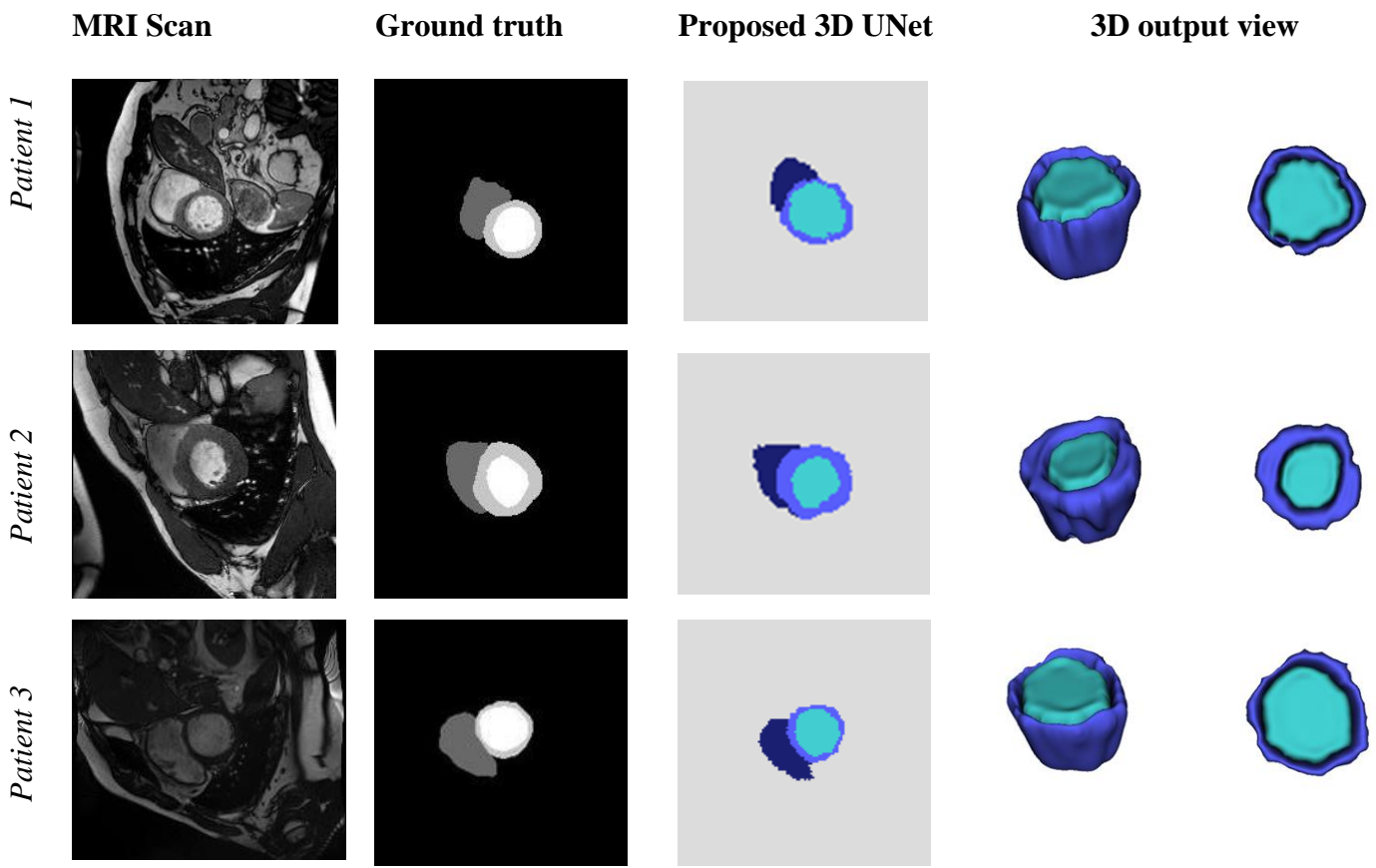

*Figure 7: Segmentation results and the ground truth mask of three Cases: In the first Three columns, input MRI, Ground truth, and exemplary test results of the segmented LV and myocardial regions extracted from MRI IMAGES. The four column illustrates the 3D view of the LV and myocardial tissues of our proposed method prediction. The LV is displayed in light blue, and the myocardium is in purple.*

The segmentation outcomes were quantitatively evaluated at two different instants of cardiac cycle: end diastolic and end systolic phases. Four metrics were used in this study: Dice Coefficient, Dice Loss, F1 score and IoU. The results of these metrics are summarized in table 5. The proposed 3D UNet achieved a DSC, Dice loss, F1 score and IoU of 0.965, 0.035, 0.801 and 0.712 on the test data of the end diastolic phase, respectively. For the end systolic phase, a DSC of 0.945 was obtained with an F1 score of 0.799 and IoU of 0.701.

**Table 5.** Quantitative assessment of our proposed model.

|  | Dice Coefficient | | Dice Loss | | F1 Score | | IoU Score | |
| --- | --- | --- | --- | --- | --- | --- | --- | --- |
| *phase instant* | ED | ES | ED | ES | ED | ES | ED | ES |
| *Training Data* | 0.958 | 0.948 | 0.042 | 0.052 | 0.818 | 0.814 | 0.703 | 0.704 |
| *Validation Data* | 0.962 | 0.955 | 0.037 | 0.045 | 0.809 | 0.805 | 0.714 | 0.709 |
| *Testing Data* | 0.965 | 0.945 | 0.035 | 0.055 | 0.801 | 0.799 | 0.712 | 0.701 |

Furthermore, the segmentation results of our proposed 3D model were compared to other methods that used the same ACDC dataset. The results show that our model outperformed the other methods and can accurately segment cardiac structures. All models were validated using both cardiac structures: LV and myocardium, as shown in table 6. These results showed that the proposed 3D model achieved higher myocardial segmentation performance in term of Dice, highlighting the significant improvement in cardiac segmentation. However, for LV segmentation, the result was very close to the one developed by Hassan et al., who used a *CondenseNet12 combined with U-Net. Compared to other architectures, the outcomes of our study and those proposed by Hassan et al. achieved the best LV segmentation results.*

**Table 6.** Performance analysis and comparison between our proposed 3D UNet and other methods using the same ACDC dataset. Best values are marked in bold font.

| *Dataset* | *Authors, years* | *Methods* | *Dice index* | |
| --- | --- | --- | --- | --- |
| | | | LV | Myocardium |
| ACDC | *Chen et al., [44], 2023* | UNet with a dilated convolution module | 0.947 | 0.899 |
| | *Zhang et al., [45], 2022* | FC-DenseNet (FCD) model and capsule convolution-capsule deconvolution | 0.925 | 0.950 |
| | *Painchaud et al., [46], 2020* | CNN | 0.933 | 0.833 |
| | *Hassan et al., [32], 2020* | CondenseNet12 and U-Net | **ED phase :0.967**<br>**ES phase :0.951** | 0.901 |
| | *Proposed method* | *Improved 3D UNet* | ED phase: *0.965*<br>ES phase :0.945 | **0.961** |

*All models were compared on the ACDC dataset using same images test (50 subjects).*

### 5.3. Clinical evaluation metrics

To study the segmentation performances of our algorithm on cardiac clinical measurements, EDV, ESV, myocardial mass and LVEF were measured using two segmentation results: with inclusion of papillary muscles and without inclusion of these structures. All the mentioned parameters were computed using the same selected images. To reduce interobserver variability, two radiologists participated in the evaluation of these measurements.

First, the clinical cardiac parameters were measured using the workstation (syngo imaging software, Siemens Healthcare, NC). LV function values with the exclusion of papillary muscles were obtained using our proposed 3D UNet model. These measurements were computed for 30 patients. The same parameters were also computed with the inclusion of papillary muscles using CVI42 software. The outcomes of both LV measurements were compared to study the impact of including or excluding papillary muscles during the segmentation process. The results are shown in Table 7 and figure 8 and are expressed as mean ± standard deviation (SD).

**Table 7**. Results of LV segmentation with/without inclusion of papillary muscles for a studied test cohort cohort (n = 30).

| *Clinical measurements* | *With inclusion of papillary muscles during segmentation* | *With exclusion of papillary muscles during segmentation* |
| --- | --- | --- |
| *Mean EDV (ml)* ± **SD** | 118 ± 44 | 122,86 ± 45 |
| *Mean ESV (ml)* ± **SD** | 50,54 ± 23 | 55,69 ± 25 |
| *Mean Myocardial mass (g)* ± **SD** | 84,95± 32 | 79,38 ± 35 |
| **Mean LVEF (%) ± SD** | 53,74 ± 14 | 50,76 ± 12 |

SD: standard deviation

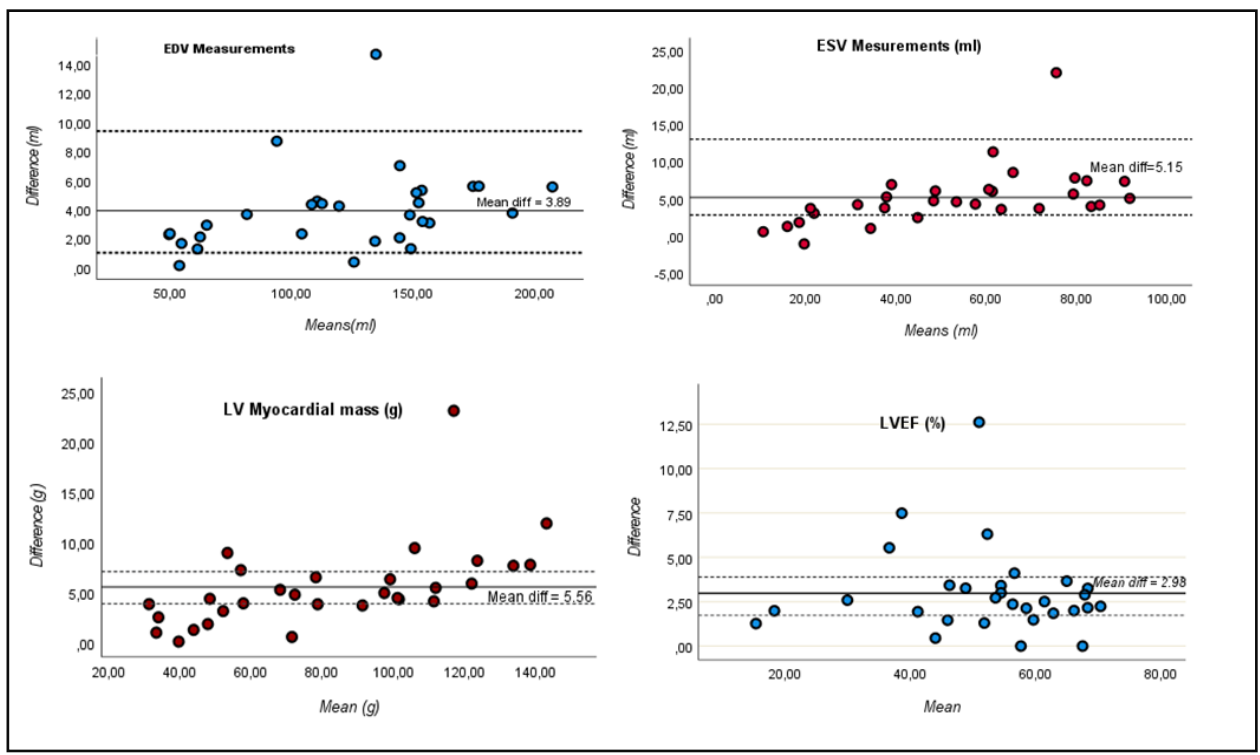

**Figure 8** Bland-Altman plots comparing clinical measurements (EDV, ESV, myocardial mass and LVEF) from LV segmentation with papillary muscles against those derived from LV segmentation without papillary muscles in the test cohort (n = 30).

The results show that LV volumes decreased using segmentation with inclusion of papillary muscles. this reduction in volume contributes to the increase of LVEF. This outcome suggests that the contribution of papillary muscles to the LVEF value is significant. Furthermore, the Bland and Altman curves showed a statistically significant difference while including or excluding papillary muscles for EDV, ESV, LVEF and myocardial mass. These findings suggest that establishing a clinical validation in addition to the segmentation validation is essential as the main objective of segmenting LV is to obtain accurate clinical measurements of LV function.

## 6. Discussion

In this study, an improved 3D UNet was proposed and validated for automated segmentation of LV and myocardium without including papillary muscles. The main outcome of our study is that the proposed 3D UNet is better suited for MRI Images since it requires only 4 slices per patient. This is a crucial point because the majority of public dataset include variation of number of frames and slices, making is difficult to apply 3D model that require a high frame numbers per patient.

 In addition, the improved 3D model showed high performances in both myocardium and LV structures with a mean dice coefficient of 0.955 and 0.961 for LV and myocardium, respectively, and an F1 score of 0.801 and 0.799, an IoU of 0.712, and 0.701 for end diastolic and end systolic phases, respectively. These metric results include different cardiac geometries related to healthy controls and subjects with various cardiac pathologies (e.g., myocarditis, infarction, …). The 3D model was also trained, tested and validated on two important instances of the cardiac cycle: end diastole and end systole phases. In both instances, good results were obtained. The table 8 provides a summary of deep learning-based methods and their performance metrics in cardiac segmentation analysis:

**Table 8:** An overview of the different deep learning models developed in literature for the detection of myocardium and LV.

| Heart Segmentation area | Authors, years | Deep learning model | Data | Performances Metrics |
|---|---|---|---|---|
| Myocardium | Romaguera et al., [47] 2018 | Deep fully convolutional neural network architecture (DCNN) | Sunnybrook Cardiac data (SCD) [48] N=420 images | Dice metric: Endo : 0.92 Epi : 093 |
| | Leclerc et al., [49] 2017 | Random Forest algorithm | 250 patients from the university Hospital of SaintEtienne | Dice metric: ED: 0.88 ES: 0.90 |
| | Kuang et al., [50] 2021 | 3D-EAR model | Myocardial Velocity Mapping Cardiac MR (MVM-CMR) [51] from18 healthy subjects N= 121 MRI images. | Dice metric: 0.89 IoU: 0.91 |
| | Kar et al., [52] 2021 | The DeepLabV3+ DCNN | the Displacement Encoding with Stimulated Echoes (DENSE) MRI sequence N= 42 images | Dice metric: 0.89 |
| | Brahim et al., [37] 2022 | UNet (ICPIU-Net) | late gadolinium enhancement magnetic resonance (LGE-MR) N= 150 images | Dice metric: 0.8765 |
| | Zou et al., [53] 2021 | local sine-wave modeling (SinMod) + U-Net | Cardiac tagged MRI dataset collected from Tongji Medical College of HUST. N = 670 | Dice coefficient: 0.882 |
| | Cai et al., [35] 2023 | 3D-Unet model | Dataset of 3D images collected from multiple centers N=116 | Dice metric: LV: 0.927 Myoc: 0.899 |
| | Yang et al., [54] 2016 | FCN architecture | The dataset includes 33 subjects N= 5,011 myocardium CMR images | Dice metric: 0.75 |
| | Jun Guo et al., [34] 2020 | 3D deeply supervised U-Net | Computed Tomography Images from 100 patients | Dice metric: 0.916 |
| | Ahmad et al., [55] 2022 | ASPP module (Atrous Spatial Pyramid Pooling) integrated with a proposed 2D-residual neural network | Dataset from56 subjects Each subject consisted of 20 frames [56] N = 1120 (56 x 20) | Dice metric: 0.854 |
| | Hasan et al, [32] 2020 | CondenseUNetis both a modification of DenseNet, as well as a combination of CondenseNet12 and U-Net | ACDC dataset [43], 150 patients | Dice metric: Myocardium: 0.901 |
| | Dhaene et al., [57] 2023 | nnU-net and Segmentation ResNet VAE | ACDC dataset [43] M&Ms dataset [58]: SCD datasets [48] N = 4578 | Dice metric: 0.828 |
| | Qiu et al., [59] 2023 | MyoPS-Net | Private dataset consisting of 50 paired multi-sequence CMR images and a public one from MICCAI2020 MyoPS Challenge [60,61] | Dice metric: 0.742 |
| | Zhu et al., [33] 2023 | 3D V-Net | MPS dataset [62], including 75 subjects. N = 1200 SPECT images | Dice metric: 0.9821 |
| | Bruns et al., [63] 2020 | CNNs 3D | two different sources, CLARITY study (18 patient) and 218 patients from the | Dice metric: (Average: 0.897) - LVM: 0.835 - RV: 0.916 |
| | | | University Medical Center Utrecht, NL | |
| | Sharkey et al.,[64], 2022 | nn-UNet | based on 1,553 patients selected from the ASPIRE registry of patients | Dice metric:<br>- LVM: 0.83<br>- RVM: 0.58 |
| | Yao et al., [65] 2023 | U-Net | CT scanner images<br>N = 68 3D images | Dice metric:<br>LVM: 0.773 |
| Left ventricle | Liu et al., [66] 2023 | CNN | The SDC dataset [67] contains 45 cine-MRI images of persons with a variety of diseases<br>N= 420 images | Dice metric: 0.944 |
| | Shoaib et al., [68] 2023 | U-net | Dataset collected from the National Heart Institute in Kuala Lumpur, Malaysia.<br>N = 6000 2D images | Dice metric: 0.9446 |
| | Kang et al., [69] 2023 | A deep neural network with an attention mechanism and a residual feature aggregation module | Dataset of Intra-Cardiac Arrest Research Using Sonography (ICARUS) | Dice metric: 0.899.<br>Precision: 0.901 |
| | Irshad et al. [40], 2024 | combination of the intelligent histogram-based image enhancement technique with a Light U-Net model | Dataset of MICCAI 2009 [70]<br>N = 7365 | Dice metric: 0.971.<br>Accuracy: 0.92 |
| | Zou et al., [53] 2021 | local sine-wave modeling (SinMod) + U-Net | Cardiac tagged MRI dataset collected from Tongji Medical College of HUST.<br>N = 670 images | Dice metric: 0.8827 |
| | Leclerc et al., [71] 2019 | CNN Techniques Based on an Encoder-Decoder Architecture | The proposed dataset consists of clinical exams from 500 patients, acquired at the University Hospital of St Etienne (France) | Dice metric: 0.95 |
| | Leclerc et al., [49] 2017 | Random Forest algorithm | University Hospital of SaintEtienne a still on-growing database of 250 patients | Dice metric:<br>- LVED: 0.92<br>- ES: 0.93 |
| | Dangi et al., [73] 2019 | U-Net +CNN | 97 patients 4D cardiac cine-MRI datasets available through the STACOM LV segmentation [74] | Dice metric: 0.849 |
| | Awasthi et al., [41] 2022 | LV network (LVNet) | Cardiac US image of the canine LV<br>N = 2262 | Dice metric:<br>WP MUSCLE: 0.851<br>WOP MUSCLE: 0.865 |
| | Shaaf et al., [75] 2022 | fully convolutional network (FCN) architecture | MRI (EMIDEC) dataset [76], 150 patients | Dice metric: 0.93 |
| | Amer et al., [77] 2021 | ResDUnet | Dataset of 2000 images acquired from 500 patients [71] | Dice metric: 0.951 |
| | Nasr-Esfahani et al., [78] 2018 | FCN | images of a database available in York university [79], contains CMRI of 33 patients, | Dice metric: 0.8724 |
| | Chhabra et al., [80] 2022 | elliptical active disc (EAD) | Statistical Atlases and Computational Modelling of the Heart (STACOM) [58]<br>N = 320 scans | Dice metric:<br>diastole phase: 0.873<br>systole phase: 0.770 |
| | Hu et al., [81] 2013 | LVOT: left ventricular outflow tract. | MICCAI [82]: three data sets obtained from the Sunnybrook Health Sciences Centre<br>N=420 images | Dice metric: 0.92. |

IoU: Intersection over Union, HD: Hausdorff Distance; AVD: absolute volume difference; RV: right ventricle; LV: left ventricular; LV MS: left ventricle myocardial scars; MYO: myocardium; WP muscle: with papillary muscle; WOP muscle: without papillary muscle, SPECT: single photon emission computed tomography.

As shown in table 8, the majority of developed study focuses on segmentation results and ignore the clinical validation step while the reproducibility of the deep learning cardiac segmentation model is dependent on the correct clinical parameters results. Furthermore, few studies have proposed a deep learning model for the segmentation of LV without including papillary muscles. Our study aligns with the latest cardiac recommendations, which suggest excluding papillary muscles from the LV volumes. In this context, clinical validation was established to highlight the impact of developing an accurate method to segment cardiac structures without including papillary muscles. The outcomes of clinical evaluation demonstrated the significance of excluding papillary muscles during segmentation of LV structure. The proposed model not only improves the segmentation results but also offers correct clinical parameters such as LVEF and myocardial mass. In fact, A significant difference was observant between the four clinical parameters when excluding or not the papillary muscles. This suggest that researchers should be aware that the presence of papillary muscles during segmentation could affect the clinical decision.

The segmentation of cardiac structures, including the myocardium, from different slices is indeed a challenging task, especially given the poor contrast often found in apical slices. Despite these challenges, our model has demonstrated good capabilities in producing accurate segmentation results across various frames and slices.

However, it is crucial to acknowledge certain limitations within proposed model. Firstly, the examination of the impact of papillary muscles during segmentation was carried out on a restricted number of patients (Only 30 patients). This number is sufficient to draw conclusions, but which we consider insufficient to announce any medical confirmation. To ensure better comprehensive, further investigations are needed to explore their influences on a more extensive patient population. In fact, the reliance of the model using only two dataset poses a risk of overfitting to that particular dataset, which could hinder generalizability to broader contexts. Addressing this concern is pivotal to enhance the model's robustness and adaptability. Actually, the performance of the model may decline when confronted with less qualities images or those containing significant artifacts. The model's efficacy could thus be compromised under such conditions, necessitating caution especially in medical image applications.

Furthermore, the fixed patch size dependency of the model and the potential introduction of artifacts due to patching are aspects that were not taken into consideration when optimizing the model's functionality and minimizing undesirable outcomes.

## *7. Conclusion*

In this study, we proposed a novel approach for cardiac MRI segmentation using a modified 3D UNet architecture. In particular, a patching technique is introduced to extract volumetric cubes from the MRI images, hence allowing the 3D UNet to learn more robust features while maintaining lower computational complexity. The proposed framework demonstrated the ability to accurately segment the left ventricle and myocardium while excluding papillary muscles, which is crucial for the precise assessment of the left ventricular function. The model achieved high Dice coefficients of 0.965 and 0.945 for the LV and the myocardium segmentation at end-diastole and end-systole phases, respectively. Additionally, the F1 scores were 0.801 and 0.799, indicating excellent overall performance.

A key strength of our approach is the ability to capture the complex 3D structure of the heart across different slices, even for the challenging apical regions with poor contrast. Moreover, the large and diverse dataset used for training, which included 5,880 images across two datasets, ensured the model's generalizability.

The exclusion of papillary muscles from the LV segmentation, as recommended by the Society for Cardiovascular Magnetic Resonance, is a key advantage of our approach. By accurately delineating the LV cavity without papillary muscles, our model enables more precise quantification of clinical parameters such as end-diastolic volume, end-systolic volume, myocardial mass, and ejection fraction. This is particularly important for accurate diagnosis and monitoring of cardiac diseases. In conclusion, the large dataset and robust 3D architecture make this approach a promising tool for automating cardiac MRI analysis and enhancing clinical decision-making.

**Acknowledgment**

The authors gratefully knowledge the staff of the Military Hospital of Instruction of Tunis, Tunisia, for their assistance in the image analysis of this study, and Dr. Arous younes for facilitating the use of imaging data.

**Declaration of competing interest**

The authors declare that they have no conflict of interest.

**CRediT authorship contribution statement**

**Narjes Benameur**: Conceptualization, Methodology, Visualization, Validation, Writing – Original Draft.: **Ramzi Mahmoudi: Methodology**, Visualization, Validation, Supervision, Review and Editing. **Mohamed Deriche**: Validation, Supervision, Writing – Review and Editing. **Amira fayouka**: Methodology and Data Curation. **Imene Masmoudi:** Methodology and Visualization. **Nessrine Zoghlami**: Validation, Supervision, Review and Editing.

**Funding sources**

This research did not receive any specific grant from funding agencies in the public, commercial, or not-for-profit sectors.

# Acknowledgments

This work was supported by Research Grant No. 2023-IRG-ENIT-40 under the Deanship of Research and Graduate Studies at Ajman University.

[54] Yang, X., Gobeawan, L., Yeo, S. Y., Tang, W. T., Wu, Z. Z., & Su, Y. (2016, septembre 14). Automatic Segmentation of Left Ventricular Myocardium by Deep Convolutional and De:convolutional Neural Networks. 2016 Computing in Cardiology Conference. https://doi.org/10.22489/CinC.2016.025-237

[55] Ahmad, I., Qayyum, A., Gupta, B. B., Alassafi, M. O., & AlGhamdi, R. A. (2022). Ensemble of 2D Residual Neural Networks Integrated with Atrous Spatial Pyramid Pooling Module for Myocardium Segmentation of Left Ventricle Cardiac MRI. Mathematics, 10(4), Article 4. https://doi.org/10.3390/math10040627

[56] Xue, W., Brahm, G., Pandey, S., Leung, S., & Li, S. (2018). Full left ventricle quantification via deep multitask relationships learning. Medical Image Analysis, 43, 54-65. https://doi.org/10.1016/j.media.2017.09.005

[57] Dhaene, A. P., Loecher, M., Wilson, A. J., & Ennis, D. B. (2023). Myocardial Segmentation of Tagged Magnetic Resonance Images with Transfer Learning Using Generative Cine-To-Tagged Dataset Transformation. Bioengineering, 10(2), Article 2. https://doi.org/10.3390/bioengineering10020166

[58] Campello, V. M., Gkontra, P., Izquierdo, C., Martín-Isla, C., Sojoudi, A., Full, P. M., Maier-Hein, K., Zhang, Y., He, Z., Ma, J., Parreño, M., Albiol, A., Kong, F., Shadden, S. C., Acero, J. C., Sundaresan, V., Saber, M., Elattar, M., Li, H., … Lekadir, K. (2021). Multi-Centre, Multi-Vendor and Multi-Disease Cardiac Segmentation : The M&Ms Challenge. IEEE Transactions on Medical Imaging, 40(12), 3543-3554. https://doi.org/10.1109/TMI.2021.3090082

[59] Qiu, J., Li, L., Wang, S., Zhang, K., Chen, Y., Yang, S., & Zhuang, X. (2023). MyoPS-Net : Myocardial pathology segmentation with flexible combination of multi-sequence CMR images. Medical Image Analysis, 84, 102694. https://doi.org/10.1016/j.media.2022.102694

[60] Zhuang, X. (2016). Multivariate Mixture Model for Cardiac Segmentation from Multi-Sequence MRI. Medical Image Computing and Computer-Assisted Intervention – MICCAI 2016: 19th International Conference, Athens, Greece, October 17-21, 2016, Proceedings, Part II, 581-588. https://doi.org/10.1007/978-3-319-46723-8_67

[61] Zhuang, X. (2019). Multivariate Mixture Model for Myocardial Segmentation Combining Multi-Source Images. IEEE Transactions on Pattern Analysis and Machine Intelligence, 41(12), 2933-2946. https://doi.org/10.1109/TPAMI.2018.2869576

[62] Weihua. (2022). MIILab-MTU/SPECTMPISeg [Logiciel]. https://github.com/MIILab-MTU/SPECTMPISeg (Édition originale 2022)

[63] Bruns, S., Wolterink, J. M., Takx, R. A. P., van Hamersvelt, R. W., Suchá, D., Viergever, M. A., Leiner, T., & Išgum, I. (2020). Deep learning from dual-energy information for whole-heart segmentation in dual-energy and single-energy non-contrast-enhanced cardiac CT. Medical Physics, 47(10), 5048-5060. https://doi.org/10.1002/mp.14451

[64] Sharkey, M. J., Taylor, J. C., Alabed, S., Dwivedi, K., Karunasaagarar, K., Johns, C. S., Rajaram, S., Garg, P., Alkhanfar, D., Metherall, P., O'Regan, D. P., Van Der Geest, R. J., Condliffe, R., Kiely, D. G., Mamalakis, M., & Swift, A. J. (2022). Fully automatic cardiac four chamber and great vessel segmentation on CT pulmonary angiography using deep learning. Frontiers in Cardiovascular Medicine, 9, 983859. https://doi.org/10.3389/fcvm.2022.983859

[65] Yao, Z., Xie, W., Zhang, J., Yuan, H., Huang, M., Shi, Y., Xu, X., & Zhuang, J. (2023). Graph matching and deep neural networks based whole heart and great vessel segmentation in congenital heart disease. Scientific Reports, 13, 7558. https://doi.org/10.1038/s41598-023-34013-1

[66] Liu, J., Yuan, G., Yang, C., Song, H., & Luo, L. (2023). An Interpretable CNN for the Segmentation of the Left Ventricle in Cardiac MRI by Real-Time Visualization. Computer Modeling in Engineering & Sciences, 135(2), 1571-1587. https://doi.org/10.32604/cmes.2022.023195

[67] Sunnybrook Cardiac Data – Cardiac Atlas Project. (s. d.). https://www.cardiacatlas.org/sunnybrook-cardiac-data/ (accessed April 6, 2024)

[68] Shoaib, M. A., Chuah, J. H., Ali, R., Dhanalakshmi, S., Hum, Y. C., Khalil, A., & Lai, K. W. (2023). Fully Automatic Left Ventricle Segmentation Using Bilateral Lightweight Deep Neural Network. Life, 13(1), Article 1. https://doi.org/10.390/life13010124

[69] Kang, S., Kim, S. J., Ahn, H. G., Cha, K.-C., & Yang, S. (2023). Left ventricle segmentation in transesophageal echocardiography images using a deep neural network. PLOS ONE, 18(1), e0280485. https://doi.org/10.1371/journal.pone.0280485

[70] Deng, J., Dong, W., Socher, R., Li, L.-J., Li, K., & Fei-Fei, L. (2009). ImageNet : A large-scale hierarchical image database. 2009 IEEE Conference on Computer Vision and Pattern Recognition, 248-255. https://doi.org/10.1109/CVPR.2009.5206848

[71] Leclerc, S., Smistad, E., Pedrosa, J., Østvik, A., Cervenansky, F., Espinosa, F., Espeland, T., Berg, E. A. R., Jodoin, P.-M., Grenier, T., Lartizien, C., D'hooge, J., Lovstakken, L., & Bernard, O. (2019). Deep Learning for Segmentation Using an Open Large-Scale Dataset in 2D Echocardiography. IEEE Transactions on Medical Imaging, 38(9), 2198-2210. https://doi.org/10.1109/TMI.2019.2900516

[72] Lindseth, F. (s. d.). Real-time Tracking of the Left Ventricle in 3D Ultrasound Using Kalman Filter and Mean Value Coordinates.
18